\documentclass{article}
\usepackage{ijcai25}

\usepackage{times}
\usepackage{soul}
\usepackage{url}
\usepackage[hidelinks]{hyperref}
\usepackage[utf8]{inputenc}
\usepackage[small]{caption}
\usepackage{graphicx}
\usepackage{amsmath}
\usepackage{amsthm}
\usepackage{booktabs}
\usepackage{algorithm}
\usepackage{algorithmic}
\usepackage[switch]{lineno}

\usepackage{listings}
\newtheorem{example}{Example}

\title{Bridging the Reasoning Gap: Small LLMs Can Plan with Generalised Strategies}

\author{
Andrey Borro
\and
Patricia J Riddle\and
Michael W Barley\And
Michael J Witbrock\\
\affiliations
University of Auckland
\emails
abor198@aucklanduni.ac.nz,
\{p.riddle, m.barley, m.witbrock\}@auckland.ac.nz
}

\begin{document}

\maketitle

\begin{abstract}
    Recent advancements in the reasoning skills of Large Language Models (LLMs) demonstrate an increase in the ability of LLMs to solve simple planning tasks. However, as long as the driving force behind improved reasoning capability is the size and complexity of the model, the financial and computational costs associated with running them will also increase. This trend raises questions about continued accessibility and whether these improvements will increase at the same pace as models continue to grow in size and expense. We propose two approaches to enhance the reasoning ability of less resource-intensive LLMs. (1) Provide them with a generalised strategy for solving tasks within a given domain, generated by a more resource-intensive LLM. (2) Exploit their cost-effectiveness by iteratively prompting these models to correct errors in their proposed solutions. Our empirical results from planning and mathematical reasoning tasks demonstrate that these methods improve the performance of less resource-intensive LLMs to levels comparable with their more resource-intensive counterparts, at a fraction of the cost. Additionally, we show that the utilisation of generalised strategies in our experiments reduced the cost of the less resource-intensive model by nearly 30 percent on average.

\end{abstract}

\section{Introduction}
The emergence of reasoning capabilities in Large Language Models (LLMs) has prompted research into their ability to solve a wide variety of problems~\cite{bubeck-gpt4experiments-reasoning,zhao-marco-o1-reasoning,yu:alert-reasoning-tasks}. Further research into prompt engineering ~\cite{prompteng-job2023} and prompting techniques such as Chain-of-Thought~\cite{wei:cot} has improved the ability of pre-trained LLMs to solve reasoning tasks without fine-tuning. Despite these advances, many LLMs still perform poorly on System 2~\cite{kahneman:systems1and2} reasoning tasks in maths, logic, and planning.

LLMs empowered with internal reasoning steps, such as OpenAI's o1~\cite{openai:o1-systemcard}, which iteratively feeds its own output back in as input for Chain-of-Thought-like reasoning, have shown promise in solving planning tasks, overcoming some limitations previously faced by LLMs. However, as evidenced by recent work~\cite{rao-o1}, these successes are restricted to small-scale tasks within simple and transparent domains. Furthermore, there is a clear correlation between the size, computational requirements, and financial cost of an LLM and its performance on planning tasks. For example, when tasked with manipulating a small handful of blocks on a table, the weaker and more affordable o1-mini can solve less than half of the tasks solved by the more powerful and expensive o1.

Progress in improving the reasoning capabilities of LLMs through the development of more resource-intensive models is beneficial. However, it is important to recognise that data is finite, and the continued returns on scaling models with increased amounts of data, greater compute power, or more parameters remain uncertain. Additionally, rising financial and computational demands render such models less accessible. This serves to widen the gap with traditional planning systems, such as FastDownward~\cite{helmert:fastdown}, which is capable of solving planning tasks with negligible operating costs and in a small fraction of the time taken by an LLM while guaranteeing solution optimality and correctness.

We propose an approach to augment the reasoning abilities of weaker models by applying the expertise of more powerful models. This is accomplished by prompting the more powerful model to generate a generalised strategy for solving tasks within a given domain. The strategy is then integrated as additional non-parameterised knowledge into the input text of the weaker model. This has a substantial effect on the average cost of solving a task in cases where the cost of generating the strategy can be amortised across several tasks. Additionally, we explore a complementary technique that further boosts model performance by identifying errors in its solutions and prompting it to make corrections accordingly. 

Our methods are capable of strongly improving the performance of weaker LLMs on reasoning tasks such as planning and maths. In our experiments, weaker models utilising our methods consistently performed at a level similar to the more powerful model used as the baseline, for a fraction of the cost. Furthermore, we show that even for the same model, the incorporation of a generalised strategy into the task prompt reduces reasoning token costs by 50 percent when solving tasks with four rounds of error correction and by 30 percent without. We demonstrate that our methods can be extended beyond planning tasks to mathematical reasoning by solving 95 percent of a challenging CRT dataset~\cite{xie:crt-modified}. This result outperforms the more powerful model used as the baseline by 20 percentage points, for less than one-tenth of the cost. 


\section{Background}
\subsection{Automated Planning}
Automated Planning is a branch of Artificial Intelligence primarily focused on the generation of plans. Plans are sequences of actions within a given environment that can be sequentially applied to an initial state. As this paper explores the ability of LLMs to generate solutions to planning tasks, the words plan and solution are used interchangeably. 

A planning task $\Pi$ is a 6-tuple $(S, s_i, S_g, A, f_a, f_t)$ consisting of: 
\begin{itemize}
    \item A finite and discrete set of all states $S$, also known as the state space.
    \item The initial state $s_i \in S$. 
    \item The set of goal states $S_g \subseteq S$.
    \item The set of actions $A$.
    \item The applicability function $f_a$, which returns $f_a(s) \subseteq A$, i.e. the set of all actions applicable at state $s$. 
    \item The transition function $f_t$, which maps a state $s$ and applicable action $a$ to the resultant state $s'$ of applying $a$ to $s$, such that $s' = f_t(s, a)$, $a \in f_a(s)$.    
\end{itemize}

In practice, most planning systems \cite{planning1970,idastar2021} perform graph search on a graph $G = (V,\, E)$, where $V$ is the set of vertices and $E$ is the set of edges. Vertices are states $s \in S$, while edges are formed from the transition function $f_t$, where 
\[
\forall\, u, v \in V, \quad (u, v) \in E \iff \exists\, a \in A \colon f_t(u, a) = v.
\]

A correct solution for a given planning task is a plan $a_1 ... a_n$ such that one can form a path $v_0 ... v_{n}$ through $G$, where $v_i = f_t(v_{i-1},\,a_i)$, $v_0 = s_i$ and $v_n = s_g \in S_g$; that is, a plan that transforms $s_i$ into $s_g$ when applied sequentially. Heuristics~\cite{heuristic1968,heuristics1982} are often used to guide the search and reduce the number of vertices that must be visited for a solution to be found. 

The Planning Domain Definition Language (PDDL)~\cite{ghallab:pddl} is a formal language that is widely used for the representation of domains and tasks in planning. In our experiments, the descriptions of planning tasks, states, domains, actions, and plans are represented in PDDL within our prompts and in the final output of the LLM. However, a full understanding of our research does not require prior knowledge of PDDL. 

\subsection{Large Language Models}
Large Language Models (LLMs)~\cite{llmcodex2021,llmeval2024,llmmultimodal2023} are very large neural networks, typically containing billions of parameters. These models are trained through unsupervised learning on vast corpora of human-written text to effectively process long text sequences in parallel and capture non-contiguous word dependencies.

Prompt engineering~\cite{prompteng-patterncatalog2023,prompteng-promptcatalog2024,shin:autoprompt} is a common approach to enhancing the reasoning performance of LLMs without fine-tuning their parameters or altering their architecture. With prompt engineering, specific prompts are crafted to influence how the model's parametrised knowledge interacts with the input text. This can involve techniques like providing worked examples, phrasing the query in a question-and-answer format, or asking the model to `think through its answer step-by-step'~\cite{kojima:step-by-step}.

\subsection{Related Work}
\paragraph{Planning with LLMs}
The versatility of LLMs has resulted in a broad intersection between LLM research and planning methodologies. Some research leverages the capability of LLMs to store specialised domain knowledge, either in their pre-trained state for general environments~\cite{zeroshot2022,progprompt2023} or through fine-tuning for specific domains~\cite{finetunegpt2}. Other studies explore planning processes augmented by external planning systems~\cite{rao:llm-modulo} or translation methods to convert complex natural-language tasks into representations compatible with existing planning and constraint satisfaction algorithms~\cite{hao:translate-planning}.

While complex systems utilising LLMs as part of the planning process have shown progress, directly passing a planning task as input and obtaining a solution as output remains challenging. However, some advances have been made in guiding LLMs to decompose planning tasks into subtasks~\cite{wang:plan-and-solve-decomposer,shen:hugginggpt-decomposer,worldmodel2023}, and increasingly large and powerful LLMs have demonstrated modest success with smaller tasks~\cite{rao-o1}.

Similar to our work, Silver et al.~\shortcite{silver:genplanning} explore the extraction of generalised knowledge from LLMs in the form of executable programs for solving specific domains. Their approach aims to circumvent the costly search processes employed by traditional planning systems by generating a tailored, non-search algorithm for task resolution within a specified domain. However, their method encounters difficulties when applied to domains with complex relationships between objects. Additionally, its effectiveness in handling challenging domains that lack structured and repeatable generalised solutions, or domains encompassing an extensive range of task configurations, remains uncertain. In contrast, our approach involves extracting generalised knowledge in the form of high-level textual prompts to guide the reasoning process of an LLM, which is articulated through fluid natural language output and demonstrates flexibility in handling diverse inputs and edge cases.

\paragraph{Empowering weaker models}
The primary method of enhancing the performance of weaker LLMs is through manipulating the pre-trained parameters of the model to better suit a specific task or domain. This is often done with synthetic data~\cite{ma:star-synthetic} or knowledge distillation~\cite{gou:knowledge-distill} from a more powerful language model.

Approaches to enhance LLM performance without affecting the model parameters often focus on manipulating the prompt through prompt engineering or methods such as RAG~\cite{lewis:rag}. Wang et al.~\shortcite{wang:customertutor} provide a knowledge base of scenario and response pairs for complex interactions as examples to aid decision-making in a customer service environment. This approach is promising, but demands an initially strong level of reasoning and analytical skill from the LLM to analyse examples. Our approach swaps out a broad collection of open-ended customer interaction examples for a single set of generalised instructions and observations about a specific domain or class of reasoning tasks. 



\begin{figure}[t]
    \centering
    \includegraphics[width=0.75\columnwidth]{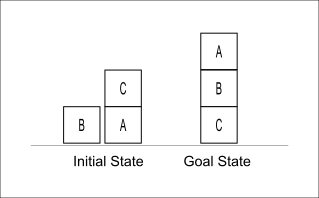}
    \caption{A simple example of a BlocksWorld initial state and goal state.}
    \label{fig:blocks_example}
\end{figure}


\section{Methodology}
\begin{figure*}[t]
    \centering
    \includegraphics[width=\linewidth]{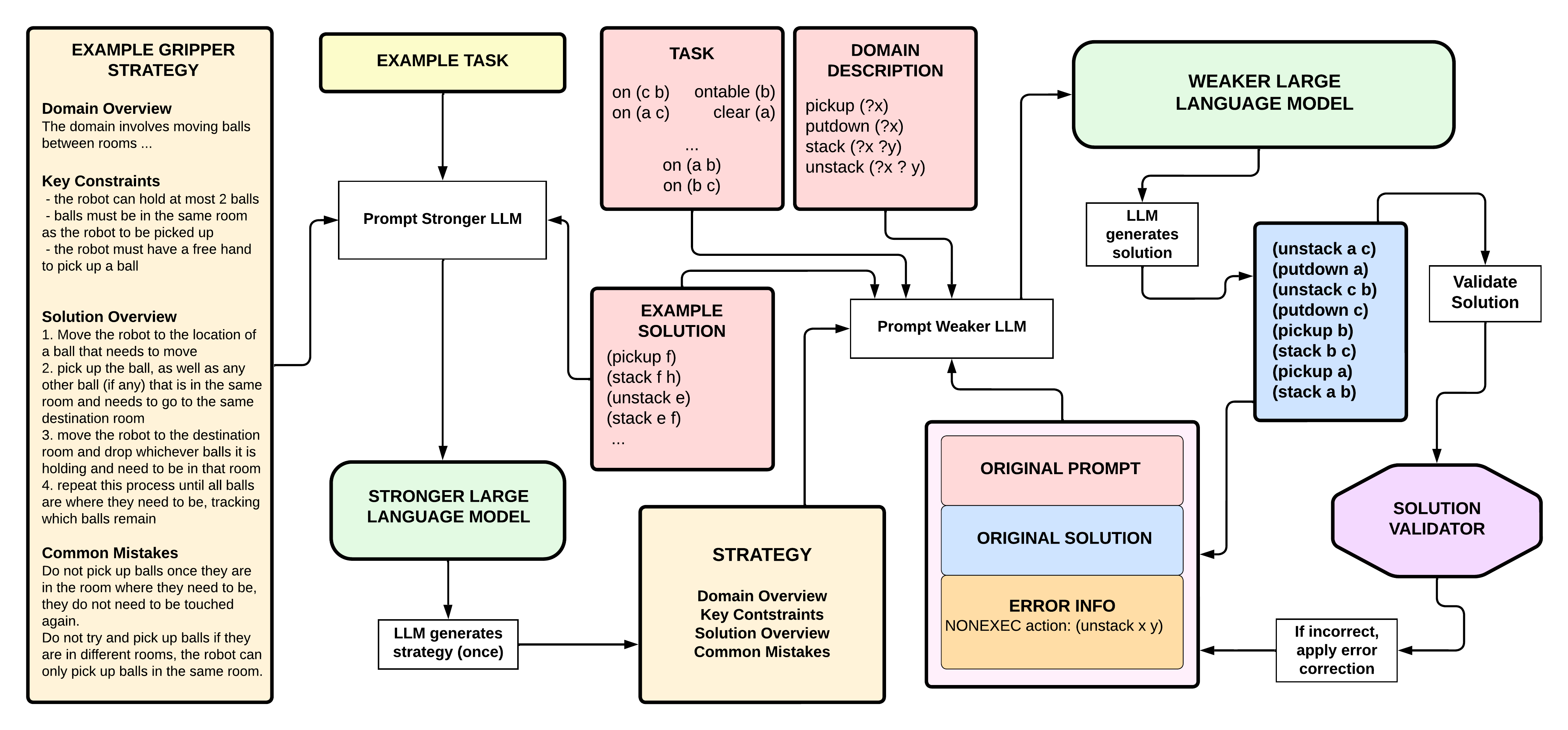}
    \caption{A diagram of our approach to enhancing the reasoning ability of weaker LLMs through generalised strategies and error correction.}
    \label{fig:diagram}
\end{figure*}

Figure~\ref{fig:diagram} presents a diagram of our proposed approach. A comprehensive set of all prompts, strategies, domain descriptions and tasks used in our paper, as well as the source code to replicate our experiments, can be found at the link in Section~\ref{section:declaration}.

\paragraph{BlocksWorld Task.} The BlocksWorld domain~\cite{slaney:blocksworld-revisited}, used in our experiments, involves manipulating configurations of blocks on a table. An example of an initial and a goal state in this domain is shown in Figure~\ref{fig:blocks_example}. 

A prompt to solve a BlocksWorld task consists of the domain description for BlocksWorld, an example solution, and the initial state and goal state for the task to be solved. With the exception of the example solution, this information is generally present in a standard PDDL description of a planning task. The expected output is a sequence of actions to solve the task. 

The example solution, provided with the intent of making the solution format clear, correctly solves a simple BlocksWorld task that is not present in the experiment dataset. The task solved by the example solution and the reasoning steps to solve it are not provided in order to avoid the LLM interpreting the example task's PDDL as part of the real task's PDDL. When providing a strategy, it is appended to the prompt between the domain and the example solution.

\paragraph{Generating a Strategy.} To generate a generalised strategy, we prompt the strategy-generating LLM with the domain description for BlocksWorld, an example BlocksWorld task and solution, a set of instructions for how to construct the main strategy body, and an example strategy for a common domain known as Gripper~\cite{alarnaouti:reformulation-gripper}, centred around moving balls between rooms. The strategy body comprises four parts: 1) a high-level description of the domain; 2) a list of key constraints within the domain; 3) a generalised set of instructions for solving a task within the domain; 4) a set of mistakes that the LLM predicts could occur while solving a task within the domain. 

\paragraph{Handwritten Strategy.} Our handwritten strategy follows the same structure as the LLM-generated strategies; that is, a high-level description of BlocksWorld, a list of key constraints, a set of generalised instructions for solving a BlocksWorld task, and some common mistakes that we predicted could arise during the solving process. For BlocksWorld, our proposed instructions follow a non-optimal but simple strategy. We propose that the LLM tasked with solving the task deconstructs all the towers to begin, laying out the blocks on the table individually. Then, we propose that it sequentially stacks the separated blocks to form the goal tower.

\paragraph{Error Correction.}
An error correction prompt is constructed from the original prompt for solving the problem, the incorrect solution, and an error message. For planning tasks, the error message includes whether or not the entire solution could be executed, the first non-executable action and the last state before failure during the execution process.

Error correction is applied exclusively to solutions identified as incorrect. In each round of error correction, only those tasks that remain unresolved after the initial round and all preceding correction rounds are re-submitted for evaluation. The LLM is provided with the incorrect solution from the last round and does not have access to information about prior attempts. 

\section{Results}
This section presents our results on the BlocksWorld planning domain and a variation of the CRT maths reasoning dataset~\cite{xie:crt-modified}. In order to improve the clarity and reproducibility of our work, we list in Table~\ref{tab:models_table} the full model numbers of all five OpenAI models used in our experiments, as well as their current API costs by input and output tokens.

\subsection{Experimental Setup}
\begin{table}[t]
\centering
\begin{tabular}{llrr}
\toprule
\textbf{Name} & \textbf{Model} & \textbf{I(\$)} & \textbf{O(\$)}\\
\midrule
o1 & \text o1-preview-2024-09-12 & 15.00 & 60.00\\
o1-mini & o1-mini-2024-09-12 & 3.00 & 12.00\\
4o & gpt-4o-2024-11-20 & 2.50 & 10.00\\
4o-mini & gpt-4o-mini-2024-07-18 & 0.15 & 0.60\\
3.5-turbo & gpt-3.5-turbo-0125 & 0.50 & 1.50\\
\bottomrule
\end{tabular}
\caption{List of OpenAI models used in this paper, and their API cost per million input (I) or output (O) tokens. Prices are in USD as of Jan 2025.}
\label{tab:models_table}
\end{table}

We evaluate the effect of four generalised strategies on the success rate for o1-mini across 50 randomly generated BlocksWorld tasks. The tasks contain 5-6 blocks and have optimal solutions of 16-18 actions. As such, they are at least as difficult as the largest tasks in the main dataset used by Valmeekam et al.~\shortcite{rao-o1}, both by number of blocks and by solution length. 

Three of the strategies are generated by o1 at test time, while one is handwritten by us in the same format as the LLM-generated strategies. As a baseline, we also evaluate the tasks on both o1 and o1-mini without providing a supplementary strategy. This results in six combinations of model and strategy, all of which undergo four rounds of error correction after an initial round of solving the tasks. Only the incorrect solutions from each round are selected for correction in the next. 

\begin{table}[t]
\centering
\begin{tabular}{lrrrrr}
\toprule
\textbf{Experiment} & \textbf{Initial} & \textbf{R1} & \textbf{R2} & \textbf{R3} & \textbf{R4}\\
\midrule
o1 & 88 & 98 & 100 & 100 & 100\\
o1-mini & 30 & 40 & 54 & 66 & 68\\
o1-mini + HS & 98 & 100 & 100 & 100 & 100\\
\midrule
o1-mini + GS1 & 56 & 76 & 90 & 92 & 94\\
o1-mini + GS2 & 50 & 66 & 76 & 86 & 90\\
o1-mini + GS3 & 32 & 60 & 70 & 82 & 86\\
\midrule
GS Average & 46.0 & 67.3 & 78.7 & 86.7 & 90.0\\
\bottomrule
\end{tabular}
\caption{Success rates (\%) on 50 BlocksWorld tasks, with four rounds of iterative error correction. We compare the o1-mini and o1 baselines to o1-mini enhanced with both handwritten (HS) and generated (GS) strategies. GS Average is the average success rate across the three generated strategies.}
\label{tab:main_table}
\end{table}

\subsection{Success Rates}
The success rates for our experiments are presented in Table~\ref{tab:main_table}. There is a large difference in the baseline performance between o1 and o1-mini on our BlocksWorld dataset, consistent with the findings outlined by Valmeekam et al.~\shortcite{rao-o1}.

\paragraph{Error Correction.} 
Error correction improves the success rate of all six experiment variations. For the baseline o1-mini and the three generated strategies, adding four steps of error correction after the initial round roughly doubles their success rates. Notably, even just one error correction step has a strong impact, increasing the average success rate for o1-mini by 22 percentage points when using generated strategies. Baseline o1 solves 88 percent of the tasks before error correction is even applied. However, the first round of error correction further increases this by 10 percentage points, demonstrating that there is no visible pattern of some BlocksWorld tasks of this size being consistently unsolvable for o1. As such, our method of error correction is an efficient way to increase success rates even for already well-performing systems.

In order to evaluate the impact of including information about the error in the error correction prompt, we extended our experiments by utilising a different technique for our error correction rounds. Instead of using a specific error prompt, at each round of error correction, we repeated the initial task prompt for any incorrect solutions. This approach achieved 62 percent for baseline o1-mini and 89 percent when averaged across the three generated strategies. 

There is no notable difference between providing detailed error feedback and simply repeating the initial prompt. This suggests that, while our specific method is not harmful, the effects of our error correction primarily stem from repeatedly attempting to solve incorrect tasks. Therefore, error correction is likely to be effective for tasks where the solution can be easily validated, even when one is not able to pinpoint exact errors. 

\paragraph{Effect of Generated Strategies.} 
Incorporating any of the three generated strategies into the task prompt substantially enhances the ability of o1-mini to solve BlocksWorld tasks. Apart from the outlier in the initial round with strategy 3, each of the three strategies consistently outperforms the baseline by at least 20 percentage points at any given round. This increase in performance is sufficient to make o1-mini competitive with the more powerful o1. After four error correction steps, applying generated strategies results in an average success rate of 90 percent, which is slightly higher than that of o1 with no strategy or error correction.

\paragraph{Handwritten.} Incorporating our handwritten strategy into the base prompt results in near-perfect success from o1-mini. In fact, in the initial round, it fails to solve only 1 task out of 50. This places the initial success rate of o1-mini with a handwritten strategy at 68 percentage points above baseline o1-mini, and 10 percentage points above baseline o1. Our results demonstrate a gap between the quality of the generalised strategies generated at test time by o1 and those written by a human with specialised knowledge of the domain. It is unclear whether this gap can be closed without improving the model tasked with generating the strategy.

\begin{table*}[t]
\centering
\begin{tabular}{lrccrrc}
\toprule
 & \multicolumn{2}{c}{Initial} & \multicolumn{4}{c}{After Error Correction}\\
 
\cmidrule(lr){1-1} \cmidrule(lr){2-3} \cmidrule(lr){4-7}

\textbf{Experiment} & \textbf{TPT} & \textbf{CPS(\$)} & \textbf{EC-TPT} & \textbf{Tokens} & \textbf{Cost(\$)} & \textbf{CPS(\$)}\\
\midrule
o1 & 5493 & 0.37 & 5440 & 312768 & 18.77 & 0.38\\
o1-mini & 6754 & 0.27 & 6367 & 1006272 & 12.08 & 0.36\\
o1-mini + HS & 2435 & 0.03 & 3136 & 124928 & 1.50 & 0.03\\
\midrule
o1-mini + GS1 & 3978 & 0.09 & 4029 & 372160 & 4.47 & 0.10\\
o1-mini + GS2 & 4958 & 0.12 & 5497 & 583296 & 7.00 & 0.16\\
o1-mini + GS3 & 5208 & 0.20 & 4430 & 606016 & 7.27 & 0.17\\
\midrule
GS Average & 4714 & 0.13 & 4652 & 520490 & 6.25 & 0.14\\
\bottomrule
\end{tabular}
\caption{Token and Cost Analysis of our experiments in Table~\ref{tab:main_table}. We report average tokens per task (TPT) during both initial and error correction (EC-TPT) rounds, total and initial cost per successfully solved task (CPS), total token usage (Tokens), and total cost (Cost) after four rounds of error correction. 
}
\label{tab:tokens_table}
\end{table*}

\subsection{Token Analysis}
OpenAI's o1 and o1-mini achieve enhanced performance on reasoning tasks through performing internal reasoning steps. In order to do this, they feed their output tokens back into themselves as input. Tokens generated during this process are referred to as reasoning tokens and are (as of January 2025) priced at the same rate as output tokens. The quantity of reasoning tokens generated by a specific model is a good metric by which to compare both the computational and financial costs of solving a task. For each experiment variation, Table~\ref{tab:tokens_table} presents: 
\begin{itemize}
    \item The average number of reasoning tokens generated per task during the initial round (TPT).
    \item The (USD) cost per solved task (CPS) in the initial round.
    \item The average number of reasoning tokens generated when performing error correction (EC-TPT).
    \item The total number of reasoning tokens generated across five rounds (Tokens).
    \item The total (USD) cost of the reasoning tokens generated (Cost).
    \item The (USD) cost per solved task (CPS) across five rounds.
\end{itemize}

\paragraph{Token Count.} In the initial round, there is an inverse correlation between the task success rate from Table~\ref{tab:main_table} and the average token usage per task for experiment variations on o1-mini. Incorporating our handwritten strategy into the task prompts decreases the average reasoning tokens per task used by o1-mini by over 4000 tokens when compared to the baseline, while incorporating the generated strategies averages around 2000 saved tokens per task. In our experiments, we identified that there was no notable difference in average token usage based on task outcome (success or failure). Consequently, the observed differences in average token usage across experiment variations cannot be attributed to a disproportionate distribution of ‘cheaper’ outcome types. Any variation in average tokens per task must arise from factors other than the relative frequencies of those outcomes.

The inverse relationship between the \textbf{total} reasoning token usage across five rounds and the success rates in Table~\ref{tab:main_table} is causal. Tasks that are not solved in a given round will require additional tokens in the next round when error correction occurs. There is no notable difference between the average tokens used by the first attempt to solve a problem and the subsequent error correction steps. After four rounds of error correction, incorporating our handwritten strategy into the task prompt reduces overall token usage by almost nine times when compared to the baseline, while incorporating generated strategies reduces overall token usage by around half.

\paragraph{Cost Analysis.} 
Our results in Table~\ref{tab:tokens_table} show that baseline o1 is the most expensive of the experiment variations. This is unsurprising, as the price of a reasoning token is five times higher for o1 than for o1-mini. As such, in the initial round, any experiment variation running on o1-mini is guaranteed to be more cost-efficient per successful task if its success rate exceeds 20 percent. Likewise, with only four rounds of error correction, any variation running on o1-mini should not cost more than o1 in total, regardless of success rate.

Despite baseline o1-mini being more cost-efficient per successful task, o1 is generally preferable due to its high success rates. The importance of this is more clearly visible in the total cost per successful task (total CPS), where the cost to run baseline o1-mini per task solved successfully increases by one-third from the initial round (initial CPS). By the end of the fourth error correction round, the total CPS for baseline o1-mini is equivalent to that of baseline o1, despite a 33 percentage point difference in success rate.

This increase in total CPS is within expectations, as tasks that were not solved in the initial round are likely to be more difficult and, therefore, require more tokens on average to solve or simply be too hard to solve altogether. In fact, the slight decrease from initial CPS to total CPS with GS3 is likely to be an outlier due to GS3's unexpectedly poor success rate in the initial round, as evidenced in Table~\ref{tab:main_table}.

Expecting an increase in CPS makes the results of including a generalised strategy in the task prompt more impactful. We note that, on average, including a strategy in the task prompt for o1-mini eliminates any notable difference between the initial CPS and total CPS. This demonstrates a linear rate of return; that is, across five rounds and up to an average 90 percent success rate, increasing the initial success rate by a factor of $n$ when using a strategy should only increase the cost by the same factor $n$.

A linear rate of return has a strong impact on the cost-efficiency of solving BlocksWorld tasks. Including a strategy in the base prompt for o1-mini and running four rounds of error correction results in a better success rate than baseline o1 for a third of the cost.  As with the success rates, the effect on cost is amplified when using the handwritten strategy. Solving all 50 tasks using o1-mini with a handwritten strategy costs less than solving just four tasks with o1. 

Even without taking error correction and the effect of success rates into account, generalised strategies increase the cost-efficiency of solving BlocksWorld tasks. The additional cost to include a generated strategy, which averages 1200 tokens in size, is only 18 cents for the 50 tasks. Comparatively, reducing the reasoning tokens per task by 2000 tokens on average is a 30 percent decrease, which saves around \$1.20 for the 50 BlocksWorld tasks on o1-mini. To generate a strategy, o1 uses 1200 output and 3200 reasoning tokens on average, costing a total of 25 cents. This means that, just by saving reasoning tokens in the initial round, a generated strategy for o1-mini will pay off after only 13 BlocksWorld tasks.

\subsection{Larger Problems}
A key limitation for LLM reasoning on BlocksWorld tasks is that model performance degrades quickly as the complexity of the task increases. Valmeekam et al.~\shortcite{rao-o1} report that, as the number of blocks within a task increases from 3-5 to 6-20 and solution length grows accordingly, the success rate of o1 decreases by a factor of four. 

In order to evaluate whether error correction and the incorporation of strategies into the base prompt could overcome this limitation, we test o1-mini on two randomly generated datasets of 20 BlocksWorld tasks each. The first dataset contains tasks with 10-12 blocks and solutions of length 30-32. The second contains tasks with 18-19 blocks and solutions of length 40-44. They are referred to as `mid' and `large' respectively. 

We evaluate our datasets on o1-mini in three variations: without a strategy, with a handwritten strategy, and with strategy 1, which was the best-performing LLM-generated strategy from our main experiments. The initial and final (after four rounds of error correction) success rates for both datasets are presented in Table~\ref{tab:scaling_table}. 

\paragraph{Results.} 
Our results show that with or without strategies, o1-mini still has difficulty with BlocksWorld tasks beyond a certain size. This is within expectations, as the motivation of the research was to explore ways to extract generalised and, therefore, amortisable knowledge from stronger models to enhance the performance of weaker models. o1 also has difficulty with problems of this size \cite{rao-o1}, so it would be surprising for a weaker model using o1's generalised strategy to show strong results. 

For mid-size tasks, incorporating a strategy into the task prompt has less impact on the success rate than error correction. There is some evidence that this is also the case for large-size tasks, but due to the low success rates and small dataset, this should be taken with a grain of salt.

Incorporating the handwritten strategy into task prompts has a high initial success rate and solves all tasks in both datasets after four rounds of error correction. This is a promising result, as it is not strongly affected by an increase in task size, unlike the LLM-generated strategy. However, without exploring a wider range of domains, it is not possible to make any claims about the effect of applying the right strategy on solving increasingly large planning tasks. For many domains, generalised solutions cannot be decomposed into simple steps in the same way as they can for BlocksWorld.

\begin{table}[t]
\centering
\begin{tabular}{lrrrr}
\toprule
 & \multicolumn{2}{c}{Mid} & \multicolumn{2}{c}{Large}\\
\cmidrule(lr){1-1} \cmidrule(lr){2-3} \cmidrule(lr){4-5}
\textbf{Strategy} & \textbf{Initial} & \textbf{R4} & \textbf{Initial} & \textbf{R4}\\
\midrule
None & 0 & 25 & 0 & 5\\
Generated 1 & 5 & 35 & 0 & 10\\
Handwritten & 85 & 100 & 70 & 100\\
\bottomrule
\end{tabular}
\caption{Success rates (\%) of o1-mini on two 20-task BlocksWorld datasets, with four rounds of iterative error correction. We compare no strategy to both a handwritten and a generated strategy.}
\label{tab:scaling_table}
\end{table}

\subsection{Critical Reasoning Test}
To demonstrate the transferability of applying generalised strategies to non-planning datasets, we evaluate our methods on the critical reasoning test (CRT) dataset~\cite{frederick:crt-base2005} used by Hegendorff et al.~\shortcite{hagendorff:crt-biases} to compare reasoning biases between humans and LLMs. Recent research~\cite{xie:crt-modified} has shown that, despite improvements in LLM performance on the CRT dataset, LLMs perform much worse when evaluated on variations of the original questions. Our dataset is composed of algebraic transformations of exponential growth (Type 3) CRT problems, which were presented as the most difficult of the variations explored by Xie et al. An example is given below.

\begin{example}
A pot of water is boiling on the stove, and with every 2X hours, the overall volume of the evaporated water quadruples. If the entire pot takes 8Y hours to evaporate completely, how long does it take for 1/64 of the pot to evaporate? X and Y are both numbers, you can use them to represent the final answer. Please think step by step.
\end{example}

There are 50 unique scenarios in the Type 3 CRT dataset provided by Hegendorff et al.~\shortcite{hagendorff:crt-biases}. We randomly generate 3 algebraic transformations for each, resulting in a dataset of 150 tasks. To generate the strategy, we query o1 with a handwritten task from outside the 50 scenarios in the original dataset and prompt it to write a generalised strategy for similar tasks. As o1 has a near-perfect performance on this dataset, we run our tasks on three weaker OpenAI models: gpt-4o, gpt-4o-mini, and gpt-3.5-turbo, where 4o is the strongest and 3.5 the weakest, to better evaluate the impact of our methods.  

As with the BlocksWorld tasks, our experiment variations consist of 1) a baseline with no strategy, 2) three strategies generated by o1 at test time, and 3) a handwritten strategy that we wrote ourselves. We perform two rounds of error correction on each variation. A method for breaking down a high-level solution into low-level and verifiable steps lies outside of the scope of this paper. As such, our error correction simply repeats a task if it is not solved correctly, with no additional error information. The success rates for our dataset are presented in Table~\ref{tab:crt_3_table}.

\begin{table}[t]
\centering
\begin{tabular}{lcccccc}
\toprule
& \multicolumn{2}{c}{4o} & \multicolumn{2}{c}{4o-mini} & \multicolumn{2}{c}{3.5-turbo}\\
\cmidrule(lr){1-1} \cmidrule(lr){2-3} \cmidrule(lr){4-5} \cmidrule(lr){6-7}
\textbf{Strategy} & \textbf{Init.} & \textbf{R2} & \textbf{Init.} & \textbf{R2} & \textbf{Init.} & \textbf{R2}\\
\midrule
None & 75 & 92 & 56 & 85 & 2 & 16\\
Handwritten & 96 & 96 & 90 & 97 & 28 & 58\\
\midrule
Generated 1 & 87 & 95 & 82 & 96 & 33 & 64\\
Generated 2 & 86 & 94 & 79 & 94 & 24 & 61\\
Generated 3 & 91 & 95 & 80 & 93 & 7 & 14\\
\midrule
GS Average & 88.2 & 94.9 & 80.7 & 94.7 & 21.8 & 46.7\\
\bottomrule
\end{tabular}
\caption{Success rates (\%) on 150 algebraic variations of Type 3 CRT tasks, with two rounds of iterative error correction. Across three OpenAI models, we compare no strategy to both handwritten and generated strategies. GS Average is the average success rate across the three generated strategies.}
\label{tab:crt_3_table}
\end{table}

\paragraph{Results.} 
Including a generated strategy in the task prompt has a strong impact on success rates for all three models, particularly the weaker 4o-mini and 3.5-turbo. There is a notably strong effect on 4o-mini, which, when utilising a strategy, outperforms baseline 4o on average and increases its initial-round success rate by 24 percentage points, compared to the 4o increase of 13. This is particularly notable as the cost of both input and output tokens for 4o-mini is a mere sixteenth of the respective token costs for 4o. 

Error correction is effective for all three models with or without an accompanying strategy. This confirms our findings from BlocksWorld that repeating incorrect tasks is an effective method of improving success rates, even without information about errors in prior solutions. However, it is important to note that we do this for a dataset with pre-written answers. It may be the case for CRT tasks, unlike planning, that the challenge of evaluating a solution is as difficult as solving the task.

The handwritten strategy also performs well, with a slightly higher success rate than the LLM-generated strategies on average. However, the impact of using a handwritten strategy over an LLM-generated strategy is not as strong for the CRT tasks as it is for BlocksWorld, demonstrating that there is not always a gap between the two strategy types. This discrepancy could arise from various factors, such as the quality of our handwritten strategy, the inherent characteristics of the domain, or the particular effectiveness of the LLM-generated strategy; however, determining the precise cause lies beyond the scope of this paper.

\section{Conclusions}
We propose two complementary methods for improving the reasoning performance of weaker LLMs. Our results show that our methods are effective and are able to vastly improve the ability of weaker LLMs to solve reasoning tasks not only in planning but also in mathematical reasoning domains. Utilising our methods, weaker LLMs are able to perform at a higher level than more powerful baseline LLMs and require only a fraction of the cost to do so. Furthermore, we show that generalised strategies can reduce the usage of reasoning tokens and, by extension, the cost of solving reasoning tasks even for identical LLMs. 

\paragraph{Limitations and Future Work}
A key limitation of our research is the small number of domains on which our methods are tested. Future research on a larger number of domains could improve the generalisability of our methods. It is also important to explore domains that might not have clear and easily generalisable solutions. 

Our error correction is only effective for tasks with solutions that are easier to validate than the task is to solve. This is always the case with planning tasks, but limits our ability to extend our method of error correction to all reasoning problems. 

Our methods show vast improvements for weaker LLMs solving small BlocksWorld tasks but fail to address the sharp decline in performance as task size increases that is present in our baseline. Future work could explore methods of merging our approach with hierarchical planning or problem decomposition, which could allow the models to solve larger tasks piece by piece.

The reduction in individual task reasoning cost resulting from the addition of a strategy into the prompt is quite promising. The generated strategies used in our paper were inexpensive to generate and, in the case of the CRT domain, only required a very simple prompt. More work in exploring this effect across a larger set of models and domains could reveal a simple method for generally reducing LLM reasoning costs for tasks within a common domain. 

\section{Declarations}
\label{section:declaration}
To enhance the clarity of our research and promote future work, we will open-source our code and data at \url{https://github.com/andrey-borro/reasoning-gap}.

\bibliographystyle{named}
\bibliography{references}

\clearpage
\onecolumn

\appendix
\section{Automatic Summary}
In order to aid the automatic evaluation of our tasks, we relied on a summarisation step for both BlocksWorld and Type 3 CRT tasks. For BlocksWorld, the summarisation step turns a large o1-mini output into a PDDL list of actions, mostly by extracting the actions from the larger body of the solution. This is done using gpt-4o and is quite trivial. Furthermore, 4o is substantially worse than o1-mini at solving PDDL, so there is little concern about this step secretly inflating success rates. 

The Type 3 CRT tasks are summarised by o1-mini, which poses a concern as it is much better at CRT tasks than the models whose answers it is summarising. We handle this by only providing o1-mini with the final formula answer. It never sees the original task or the reasoning behind it, and the summarisation step is posed as an algebraic reformulation task. An example of this process can be found in the LLM Output section of this Appendix (\ref{summ1} and \ref{summ2}). 

In our view, both summarisation steps worked as expected, ensuring that correct answers were marked as correct and incorrect answers were marked as incorrect. 

\section{Prompts}
\subsection{BlocksWorld Task Prompt}
\begin{lstlisting}
Consider the following PDDL domain with 4 operators:
action : pickup (block)
preconds : (clear block), (ontable block), (handempty)
effects : (holding block), not(ontable block), not(clear block), not(handempty)

action : putdown (block)
preconds : (holding block)
effects : not(holding block), (clear block), (ontable block), (handempty)\\

action : unstack (top-block bottom-block)
preconds : (on top-block bottom-block), (clear top-block), (handempty)
effects : (holding top-block), (clear bottom-block), not(on top-block bottom-block), not(clear bottom-block), not(handempty)

action : stack (top-block bottom-block)
preconds : (holding top-block), (clear bottom-block)
effects : (on top-block bottom-block), (clear top-block), (handempty), not(holding top-block), not(clear bottom-block)

[STRATEGY GOES HERE]

An example solution (without reasoning steps) would look like:
(unstack d c)
(putdown d)
(pickup c)
(stack c d)
(unstack a b)
(putdown a)
(pickup b)
(stack b c)
(pickup a)
(stack a b)

*THIS IS YOUR PROBLEM, SOLVE IT*:
Initial State:
(handempty)
(clear a)
(clear d)
(clear b)
(on d f)
(on f c)
(on b e)
(ontable a)
(ontable c)
(ontable e)

Goal:
(on a b)
(on b c)
(on c d)
(on d e)
(on e f)
Solution:
\end{lstlisting}

\subsection{Type 3 CRT Task Prompt}
\begin{lstlisting}
Solve the following problem. Give a formula enclosed in double at symbols as your final answer at the end, e.g. @@ANSWER_HERE@@

After a switch is flicked, the area of light shining within a room triples every 5X nanoseconds until it fills the room. If it takes 7Y nanoseconds for the room to fill,  how many nanoseconds after the switch was flicked was the room at 1/9 full?

A general strategy for this problem type has been given below.

[STRATEGY HERE]
\end{lstlisting}

\subsection{BlocksWorld Strategy-Generating Prompt}
\begin{lstlisting}
You will be given a PDDL domain, as well as an example problem and solution for that domain. Your goal is to write a strategy for the domain which will be used as part of a prompt for a weaker LLM to solve this problem. The strategy should be threefold. 

---
Firstly, a brief overview of the high-level dynamics, e.g. how the actions interact and what they are used for in the domain. Do not just repeat the preconditions and effects in PDDL - the LLM will have access to these. Instead, use English to explain the environment and how the actions are used to do things in it. Finish this overview with a brief list of key constraints about objects within this environment (see example).

Secondly, a high-level guide to solving a general problem within this domain. The guide should open with "To solve the problem, follow these step-by-step instructions:". Follow this with an instruction step outlining how to interpret the goal configuration in the context of planning your solution. After this, give simple and clear instructions on how to solve the task, step-by-step. Remember that you are talking to an LLM. It has no memory or space to "visualise" a problem, nor does it respond well to being philosophical. Instead, give clear and very simple actionable instructions for it to follow. Each instruction should represent a "big picture" step for the agent to follow. Then, explain how to achieve that step in sub-instructions. Absolutely prioritise the most simple (to understand and follow) instructions, even if it means having a longer solution. Do not forget that the LLM executing this is weak, and should not try to do complex reasoning if it can be avoided. Your job is to do as much of the thinking as you can, and to give it clear steps to take action on.

Thirdly, give a list (at most five) of common mistakes that you assume a weaker LLM is likely to make and how to avoid them. Prioritise the ones that you think might be most likely to stump the LLM. 

Throughout the strategy, feel free to use general examples (but keep them very short and concise, as they should not take up much of the strategy at all).

---

A (slightly less detailed than you should aim for) example for the gripper domain could look like:

Domain Overview:
The domain consists of balls (e.g. balla, ballb, ballc) in rooms (rooma, roomb ....), with a robot that travels between the two rooms. The robot can only pick up to two balls at a time using the pickup action - one in each hand, specified as part of the action e.g. (pickup balla rooma left), and only when they are with him in the room that he is currently on. The robot moves between rooms using the move action, taking the balls in his hands with him, and dropping a ball will drop it into the room that the robot is currently in. 

Key Constraints
 - the robot can hold at most 2 balls
 - balls must be in the same room as the robot to be picked up
 - the robot must have a free hand to pick up a ball

Solution Overview:
To solve the problem, follow these step-by-step instructions:
1. Observe the required goal location for the balls e.g. (at roomb balla)
2. Move the robot to the location of a ball that needs to move
- ensure that it has dropped off whichever balls it was moving prior
3. pick up the ball, as well as any other ball (if any) that is in the same room and needs to go to the same destination room
- (short example)
4. move the robot to the destination room and drop whichever balls it is holding and need to be in that room
5. repeat this process until all balls are where they need to be, tracking which balls remain

Common Mistakes:
Do not pick up balls once they are in the room where they need to be, they do not need to be touched again. 
Do not try and pick up balls if they are in different rooms, the robot can only pick up balls in the same room. 
etc. 

---

The PDDL domain that you are writing the strategy for:

action : pickup (block)
preconds : (clear block), (ontable block), (handempty)
effects : (holding block), not(ontable block), not(clear block), not(handempty)\\

action : putdown (block)
preconds : (holding block)
effects : not(holding block), (clear block), (ontable block), (handempty)\\

action : unstack (top-block bottom-block)
preconds : (on top-block bottom-block), (clear top-block), (handempty)
effects : (holding top-block), (clear bottom-block), not(on top-block bottom-block), not(clear bottom-block), not(handempty)\\

action : stack (top-block bottom-block)
preconds : (holding top-block), (clear bottom-block)
effects : (on top-block bottom-block), (clear top-block), (handempty), not(holding top-block), not(clear bottom-block)\\

Here is an example problem within this domain:

Initial State:
(clear a)
(clear d)
(handempty)
(on a b)
(on d c)
(ontable b)
(ontable c)

Goal:
(on a b)
(on b c)
(on c d)

Solution:
(unstack d c)
(putdown d)
(pickup c)
(stack c d)
(unstack a b)
(putdown a)
(pickup b)
(stack b c)
(pickup a)
(stack a b)
\end{lstlisting}

\subsection{Type 3 CRT Strategy-Generating Prompt}
\begin{lstlisting}
Take the below problem:

After a switch is flicked, the area of light shining within a room triples every 5X nanoseconds until it fills the room. If it takes 7Y nanoseconds for the room to fill,  how many nanoseconds after the switch was flicked was the room at 1/9 full?

I want you to write an explanation for a weaker LLM on how to solve similar problems. Note that the problems will have different numbers, thematic settings or even twists on algebraic variables, so make your strategy generalised.
\end{lstlisting}

\clearpage
\section{Strategies}
\subsection{BlocksWorld Handwritten Strategy}
\begin{lstlisting}
**Domain Overview:**

In this domain, you are a robot arm tasked with stacking and unstacking blocks on a table. 
You are capable of four actions, and can hold at most one block at a time.

1. (unstack x y) which unstacks block x from block y, and holds it in your hand
2. (stack x y) which stacks the block you are currently holding (x) onto a block that is at the top of a stack of blocks (y).
3. (pickup x) which picks up a block from the table and holds it in your hand.
4. (putdown x) which places the block in your hand down onto the table. 

unstack and stack are used when taking a block off a stack of blocks or putting it onto one. 
pickup and putdown are used when taking a block directly off the table or putting it directly on. 

**Key Constraints:**
- You cannot pick up or unstack any block that has a block on top if it. It must be at the top of its stack, or by itself on the table.
- You can only put a block onto the table or onto the top of a stack. It is impossible to place it down onto a block that already has a block on top.
- You cannnot pick up or unstack a block if you already have one in your hand. You hand only has capacity for one block. 
- You cannot put down a block or stack a block that you are not holding. It must be the one in your hand. 

**Solution Overview**

1. Unstack all of the blocks onto the table. 
   1.1. Pick any stack that isnt just one block high (e.g. if block x is by itself on the table, you can ignore it).
   1.2. Unstack the top block and place it onto the table
   1.3. Repeat from step [1.2] until you are down to the bottom block of the stack (the whole stack is now just on the table)
   1.4. Repeat from step [1.1] until you have done this for every stack (all the blocks are by themselves on the table)

2. Assemble the goal stack
   2.1. Pickup the next block to go on the final stack (remember that the bottom block is already on the table if you did step 1 correctly).
   2.2. Stack that block onto the stack
   2.3. Repeat from step [2.1] until complete. 
   Example: (Goal stack is `a on b on c`), then you pickup b, stack on c, pickup a, stack on b

Make sure to annotate your steps as you go, e.g. 
 - Unstacking all blocks in the stack x-z (x on the bottom, z on the top)
   - (unstack z y)
   - (unstack y x)

**Common Mistakes:**
1. **Attempting to `pickup` a Block Not on the Table**:
   - *Mistake*: Using `(pickup block)` on a block that is on another block.
   - *Solution*: Use `(unstack block block-below)` instead to remove it from the stack.

2. **Stacking onto a Block That is Not Clear**:
   - *Mistake*: Trying to stack a block onto another block that has a block on top of it.
   - *Solution*: Ensure the bottom block is clear by unstacking any blocks on top of it before stacking.

3. **Forgetting to Clear Obstructing Blocks**:
   - *Mistake*: Ignoring blocks that are on top of the blocks you need.
   - *Solution*: Unstack and put down any obstructing blocks to access the necessary blocks.

4. **Not Keeping Track of the Robot's Hand State**:
   - *Mistake*: Trying to pick up or unstack a block when the robot's hand is not empty.
   - *Solution*: Always ensure the robot's hand is empty before attempting to hold another block.

5. **Misordering the Stack Assembly**:
   - *Mistake*: Stacking blocks in the wrong order, leading to incorrect configurations.
   - *Solution*: Build stacks from the bottom up, following the exact order specified in the goal configuration.
\end{lstlisting}

\subsection{BlocksWorld Generated Strategy 1}
\begin{lstlisting}
Domain Overview:

In this domain, we work with a set of blocks that can be stacked on top of each other or placed on a table. The goal is to rearrange the blocks to match a specific configuration. You can manipulate the blocks one at a time using a robotic arm (the agent) that can hold only one block at a time.

Available actions to manipulate the blocks:

1. **Pickup (block)**: Pick up a clear block from the table when the agent's hand is empty.
2. **Putdown (block)**: Place a block the agent is holding onto the table.
3. **Unstack (top-block, bottom-block)**: Remove a clear top block from a stack when the agent's hand is empty.
4. **Stack (top-block, bottom-block)**: Place a block the agent is holding onto a clear bottom block to form a stack.

Key Constraints:

- The agent can hold only one block at a time.
- A block must be clear (no blocks on top of it) to be picked up, unstacked, or have another block stacked onto it.
- The agent's hand must be empty to pick up or unstack a block.
- To stack a block onto another, the block being stacked onto must be clear.

Solution Overview:

To solve the problem, follow these step-by-step instructions:

1. **Understand the Goal Configuration**:
   - **Identify** each block and determine where it needs to be in the final arrangement.
   - **Note** the desired stacking order from bottom to top as specified in the goal.

2. **Plan the Stacking Sequence from Bottom to Top**:
   - **List** the blocks in the order they need to be stacked, starting with the bottom block.
   - **Ensure** you know which blocks need to be moved and which are already in the correct position.

3. **Prepare the Bottom Block(s)**:
   - For each bottom block in the desired stack:
     - a. **Locate** the block.
     - b. If the block is not on the table:
        - **Unstack** it from any block below using the `unstack` action.
        - **Put it down** on the table using the `putdown` action.
     - c. **Make sure** the block is clear by removing any blocks on top:
        - If there are blocks on top, **unstack** them and **put them down** on the table.

4. **Stack Each Block in Order**:
   - For each subsequent block in your stacking sequence:
     - a. **Ensure the block to move is clear**:
        - If there are blocks on top, **unstack** them and **put them down** on the table.
     - b. **Ensure the target block (the one to stack onto) is clear**:
        - If it is not clear, **unstack** any blocks on top and **put them down** on the table.
     - c. **Pick up the block**:
        - If the block is on the table and clear, use the `pickup` action.
        - If the block is on another block and clear, use the `unstack` action.
     - d. **Stack the block onto the target block** using the `stack` action.

5. **Repeat Until All Blocks Are Stacked Correctly**:
   - **Continue** stacking each block in order, following steps 4a to 4d for each one.
   - **Verify** after each action that the blocks are in the correct intermediate configuration towards the goal.

6. **Check That the Goal State Is Achieved**:
   - **Compare** the current configuration of blocks with the goal configuration.
   - **Confirm** that each block is in the correct position and order.

Common Mistakes:

1. **Not Clearing Blocks Before Moving Them**:
   - Always ensure a block is clear (no blocks on top) before attempting to pick it up or unstack it.

2. **Forgetting to Clear the Target Block Before Stacking Onto It**:
   - Make sure the block you want to stack onto is clear before stacking.

3. **Not Emptying the Agent's Hand Before Picking Up a New Block**:
   - Remember that the agent must have an empty hand to pick up or unstack a block.

4. **Stacking Blocks in the Wrong Order**:
   - Follow the planned stacking sequence strictly from bottom to top to avoid misplacement.

5. **Ignoring Blocks Not Involved in the Final Stack**:
   - Move unnecessary blocks out of the way by placing them on the table so they don't interfere with stacking.
\end{lstlisting}

\subsection{Type 3 CRT Handwritten Strategy}
\begin{lstlisting}
**GENERAL STRATEGY**
1. Find the proportion of the full amount that your target population is at. Sometimes this is given e.g. '1/25th', but sometimes you have to work it out e.g. '5 / 45 = 1/9th'

2. Figure out how many time steps ago the target population occurred at. If the population is tripling, then 1/3rd happens one time step ago, 1/9th happens two, 1/27th happens three and so on.

3. Give the final answer. It should be TIME\_TO\_FULL - TIME\_STEPS * TIME\_STEP\_SIZE

For example, if you have: "Population takes 12M days to reach full, tripling every 5N days, how long does it take to get to 1/9th of full", then the final answer will be 12M - 2 * 5N = 12M - 10N.
\end{lstlisting}

\subsection{Type 3 CRT Generated Strategy 1}
\begin{lstlisting}
**Solving Exponential Growth Problems: A Generalized Strategy**

Exponential growth problems often involve quantities that increase (or decrease) by a constant factor over equal time intervals. These problems can vary in context, ranging from biological populations to spreading light in a room, but they share a common mathematical foundation. Here's a step-by-step guide to solving such problems, which can be applied regardless of the specific numbers or scenarios involved.

### **1. Understand the Nature of Exponential Growth**

- **Definition**: Exponential growth occurs when a quantity increases by the same factor in each equal time period.
- **Growth Factor**: The multiplier that represents how much the quantity increases over one time interval. For example, "triples" means a growth factor of 3.

---

### **2. Identify the Key Variables**

- **Growth Factor (\( r \))**: The factor by which the quantity increases each time period.
- **Time Interval (\( T \))**: The duration of one period over which the growth factor applies.
- **Total Time (\( t_{\text{total}} \))**: The total elapsed time from the start until a certain event occurs (e.g., the room is filled).
- **Desired Fraction of Completion (\( F \))**: The fraction of the total quantity you're interested in (e.g., \( \frac{1}{9} \) full).

---

### **3. Set Up the Exponential Growth Equation**

The general form of the exponential growth equation is:

N = N_0 \times r^{n}

Where:

- \( N \) = the quantity at time \( t \)
- \( N_0 \) = the initial quantity at time \( t = 0 \)
- \( r \) = growth factor per time interval
- \( n \) = number of time intervals elapsed (\( n = \frac{t}{T} \))

### **4. Relate the Quantities Using Ratios**

Since exponential growth is multiplicative, it's often useful to consider ratios:

\frac{N}{N_{\text{full}}} = \left( \frac{r^{n}}{r^{n_{\text{full}}}} \right) = r^{n - n_{\text{full}}}

Where:

- \( N_{\text{full}} \) = total quantity when complete (e.g., full room)
- \( n_{\text{full}} \) = number of intervals needed to reach \( N_{\text{full}} \)

---

### **5. Calculate the Number of Time Intervals**

- **To Reach Full Completion**:

  n_{\text{full}} = \frac{t_{\text{total}}}{T}

- **To Reach the Desired Fraction**:

  You need to find \( n \) such that:

  \frac{N}{N_{\text{full}}} = r^{n - n_{\text{full}}}

  Solving for \( n \):

  n = n_{\text{full}} + \log_{r}\left( \frac{N}{N_{\text{full}}} \right)

### **6. Solve for the Desired Time**

- **Calculate \( n \)** using the fraction \( F \):

  n = n_{\text{full}} + \log_{r}(F)

- **Find the Time (\( t \))**:

  t = n \times T

### **7. Example Application**

**Problem Statement**:
A quantity triples every \( T \) time units. It takes \( t_{\text{total}} \) time units to reach full completion. After how much time was the quantity at \( \frac{1}{9} \) of its full amount?

**Steps**:

1. **Calculate \( n_{\text{full}} \)**:

   n_{\text{full}} = \frac{t_{\text{total}}}{T}

2. **Determine \( n \) for \( \frac{1}{9} \) full**:

   Since \( \frac{1}{9} = \frac{1}{3^2} = 3^{-2} \), we have:

   \frac{N}{N_{\text{full}}} = r^{n - n_{\text{full}}} = 3^{n - n_{\text{full}}} = 3^{-2}

   Therefore:

   n - n_{\text{full}} = -2
   n = n_{\text{full}} - 2

3. **Calculate the Time (\( t \))**:

   t = n \times T = (n_{\text{full}} - 2) \times T

   Convert \( n_{\text{full}} \) back to time:

   t = \left( \frac{t_{\text{total}}}{T} - 2 \right) \times T = t_{\text{total}} - 2T

### **8. General Solution**

The time when the quantity is at a fraction \( F \) of its full amount is:

t = t_{\text{total}} + T \times \log_{r}(F)

- **Note**: Since \( \log_{r}(F) \) is negative when \( F < 1 \), this correctly subtracts time from \( t_{\text{total}} \).

### **9. Tips for Solving Similar Problems**

- **Identify the Growth Pattern**: Determine if the problem involves exponential growth or decay.
- **Express Fractions as Powers of the Growth Factor**: This simplifies calculations using logarithms.
- **Use Logarithms for Non-Integer Exponents**: When the exponent isn't a whole number, logarithms are essential.
- **Check Units Consistency**: Ensure time units are consistent throughout the calculations.
- **Interpret Negative Time Intervals**: A negative \( \log_{r}(F) \) indicates looking back in time from a known point.

 ### **10. Practice with Variations**

Try applying this strategy to different scenarios:

- A bacteria culture doubles every 3 hours and reaches a population of \( P \) in 24 hours. When was the population at \( \frac{P}{8} \)?
- A radioactive substance decays to half its mass every 5 days. How long until it reaches \( \frac{1}{16} \) of its original mass?

---

**Remember**: The key to solving exponential growth problems is understanding how quantities scale over time and using logarithms to solve for variables in the exponents. This approach works across various contexts and numerical values.
\end{lstlisting}

\clearpage
\section{LLM Output}
\subsection{Example o1-mini output to BlocksWorld No Strategy prompt}
\begin{lstlisting}
Solution:

1. `(unstack d f)`
2. `(putdown d)`
3. `(unstack f c)`
4. `(putdown f)`
5. `(unstack b e)`
6. `(stack b c)`
7. `(pickup e)`
8. `(stack e f)`
9. `(pickup d)`
10. `(stack d e)`
11. `(unstack b c)`
12. `(pickup c)`
13. `(stack c d)`
14. `(pickup b)`
15. `(stack b c)`
16. `(pickup a)`
17. `(stack a b)`

**Explanation of Steps:**

1. **Unstack `d` from `f`:** Remove block `d` from `f` to clear `f`.\\
2. **Put down `d`:** Place `d` on the table.\\
3. **Unstack `f` from `c`:** Remove block `f` from `c` to clear `c`.\\
4. **Put down `f`:** Place `f` on the table.\\
5. **Unstack `b` from `e`:** Remove block `b` from `e` to clear `e`.\\
6. **Stack `b` on `c`:** Place `b` on top of `c`.\\
7. **Pickup `e`:** Pick up block `e` from the table.\\
8. **Stack `e` on `f`:** Place `e` on top of `f`.\\
9. **Pickup `d`:** Pick up block `d` from the table.\\
10. **Stack `d` on `e`:** Place `d` on top of `e`.\\
11. **Unstack `b` from `c`:** Remove block `b` from `c` to clear `c`.\\
12. **Pickup `c`:** Pick up block `c` from the table.\\
13. **Stack `c` on `d`:** Place `c` on top of `d`.\\
14. **Pickup `b`:** Pick up block `b` from `c`.\\
15. **Stack `b` on `c`:** Place `b` back on top of `c`.\\
16. **Pickup `a`:** Pick up block `a` from the table.\\
17. **Stack `a` on `b`:** Place `a` on top of `b`.\\

This sequence of actions rearranges the blocks to achieve the goal state:

- `(on a b)`
- `(on b c)`
- `(on c d)`
- `(on d e)`
- `(on e f)`
- `(on e f)`

All blocks are stacked in the desired order with each block supporting the one above it.
\end{lstlisting}

\subsection{Example o1-mini output to BlocksWorld Generated Strategy 1 prompt}
\begin{lstlisting}
**Solution:**

To achieve the goal configuration from the given initial state, follow the step-by-step actions below. Each action manipulates the blocks to progressively build the desired stack from bottom to top.

1. **Unstack and Put Down Block `d` from `f`**:
   - **Action:** `(unstack d f)`
   - **Action:** `(putdown d)`
   
2. **Unstack and Put Down Block `f` from `c`**:
   - **Action:** `(unstack f c)`
   - **Action:** `(putdown f)`
   
3. **Unstack and Put Down Block `b` from `e`**:
   - **Action:** `(unstack b e)`
   - **Action:** `(putdown b)`

4. **Pick Up Block `e` and Stack It on `f`**:
   - **Action:** `(pickup e)`
   - **Action:** `(stack e f)`
   
5. **Pick Up Block `d` and Stack It on `e`**:
   - **Action:** `(pickup d)`
   - **Action:** `(stack d e)`
   
6. **Pick Up Block `c` and Stack It on `d`**:
   - **Action:** `(pickup c)`
   - **Action:** `(stack c d)`
   
7. **Pick Up Block `b` and Stack It on `c`**:
   - **Action:** `(pickup b)`
   - **Action:** `(stack b c)`
   
8. **Pick Up Block `a` and Stack It on `b`**:
   - **Action:** `(pickup a)`
   - **Action:** `(stack a b)`

**Complete Action Sequence:**

```
(unstack d f)
(putdown d)
(unstack f c)
(putdown f)
(unstack b e)
(putdown b)
(pickup e)
(stack e f)
(pickup d)
(stack d e)
(pickup c)
(stack c d)
(pickup b)
(stack b c)
(pickup a)
(stack a b)
```

**Explanation of the Action Sequence:**

1. **Clearing Blocks:**
   - **Unstack `d` from `f`** and **put it down** to make `f` clear.
   - **Unstack `f` from `c`** and **put it down** to place `f` on the table.
   - **Unstack `b` from `e`** and **put it down** to make `e` clear.

2. **Building the Stack from Bottom to Top:**
   - **Pick up `e`** and **stack it on `f`** to start building the stack.
   - **Pick up `d`** and **stack it on `e`**.
   - **Pick up `c`** and **stack it on `d`**.
   - **Pick up `b`** and **stack it on `c`**.
   - **Pick up `a`** and **stack it on `b`** to complete the goal configuration.

This sequence ensures that all blocks are moved in the correct order, maintaining the constraints of the domain, such as keeping the agent's hand empty before picking up a new block and ensuring that both the block to be moved and the target block are clear before stacking.
\end{lstlisting}

\subsection{Example o1-mini output to BlocksWorld Error-Correction prompt}
\begin{lstlisting}
Apologies for the oversight in the initial solution. The error occurred because the action `(stack e f)` was attempted when `(clear f)` was **not** true. In the initial state, block `d` is on `f` (`(on d f)`), making `f` **not clear**. Therefore, stacking `e` onto `f` is **not executable** at that point.

**Identified Error:**
- **Attempted to stack `e` onto `f` without ensuring that `f` is clear.** Since `d` is on `f`, `f` is not clear, violating the precondition for the `stack` action.

**Corrected Solution:**

To rectify this, we need to **first clear `f`** by removing `d` from it. However, since `a` is on `d` (`(on a d)`), we'll need to clear `a` first by removing `b` from it, and subsequently clear `b` by placing it on the table. Here's the step-by-step corrected solution:

1. **Unstack `e` from `b`:**
   (unstack e b)
   
2. **Put down `e` on the table to free the agent's hand:**
   (putdown e)
     
3. **Unstack `b` from `a`:** 
   (unstack b a)
     
4. **Put down `b` on the table:** 
   (putdown b)
     
5. **Unstack `a` from `d`:**  
   (unstack a d)
    
6. **Put down `a` on the table:**  
   (putdown a)
    
7. **Pick up `e` from the table:**
   (pickup e)
   
8. **Stack `e` onto `f`:**
   (stack e f)
    
9. **Pick up `d` from the table (since `f` is now clear):** 
   (pickup d)
    
10. **Stack `d` onto `e`:**  
    (stack d e)
      
11. **Pick up `c` from the table:**  
    (pickup c)
     
12. **Stack `c` onto `d`:**  
    (stack c d)
      
13. **Pick up `b` from the table:** 
    (pickup b)
    
14. **Stack `b` onto `c`:** 
    (stack b c)
      
15. **Pick up `a` from the table:**  
    (pickup a)
      
16. **Stack `a` onto `b`:**
    (stack a b)
    
**Complete Corrected Solution:**

(unstack e b)
(putdown e)
(unstack b a)
(putdown b)
(unstack a d)
(putdown a)
(pickup e)
(stack e f)
(pickup d)
(stack d e)
(pickup c)
(stack c d)
(pickup b)
(stack b c)
(pickup a)
(stack a b)


**Explanation of the Corrected Steps:**

1. **Clear `b` and `a` to eventually free `d`:**
   - By unstacking and putting down `e`, `b`, and `a`, we ensure that `d` becomes clear.
   
2. **Make `f` clear by moving `d` off of it:**
   - Once `a` is on the table, `d` is clear and can be unstacked from `f`.
   
3. **Reassemble the stack in the desired order:**
   - Stack `e` onto `f`, followed by `d` onto `e`, `c` onto `d`, `b` onto `c`, and finally `a` onto `b`, achieving the goal configuration.

**Verification of Goal State:**

After executing the corrected solution, the blocks will be arranged as follows:

(on a b)
(on b c)
(on c d)
(on d e)
(on e f)
(on f table)

This matches the desired goal configuration.
\end{lstlisting}

\subsection{Example gpt-4o output to Type 3 CRT No Strategy prompt}
\begin{lstlisting}
To solve the problem, let's first understand the growth process of the light area in the room:

1. The area triples every 5X nanoseconds.
2. The room is full in 7Y nanoseconds.

We need to find when the room was 1/9 full. Let's denote the full area of the room as \( A \).

**Step-by-step solution:**

- Let the initial area after the switch is flicked be \( A_0 \).
- The area grows as \( A_n = A_0 \times 3^{n} \) after each 5X nanosecond interval.
- When the room is full, we have \( A_n = A \).

Since it takes 7Y nanoseconds to fill the room, and each growth cycle is 5X nanoseconds, we first calculate the number of growth cycles:

n = \frac{7Y}{5X}

The room is full when:
A_0 \times 3^{n} = A

We want the light to be \( \frac{1}{9} \times A \):
A_0 \times 3^{m} = \frac{1}{9} \times A

Here, \( m \) is the number of cycles after which the room is 1/9 full.

Substitute \( A = A_0 \times 3^{n} \) into the equation:

A_0 \times 3^{m} = \frac{1}{9} \times (A_0 \times 3^{n})

This simplifies to:

3^{m} = \frac{1}{9} \times 3^{n}

Convert \( \frac{1}{9} \) to powers of 3:

\frac{1}{9} = 3^{-2}

Thus:

3^{m} = 3^{-2} \times 3^{n}

Using the properties of exponents, we find:

3^{m} = 3^{n - 2}

Therefore:

m = n - 2

Since each cycle is 5X nanoseconds, it means:

- The number of cycles for 1/9 full is \( n - 2 \).

Calculate the time taken:

\text{Time} = 5X \times (n - 2)

Substitute \( n = \frac{7Y}{5X} \):

\text{Time} = 5X \times \left(\frac{7Y}{5X} - 2\right)

Simplify:

\text{Time} = 7Y - 10X

So, the number of nanoseconds after the switch was flicked when the room was 1/9 full is:

\boxed{7Y - 10X}

Thus, the final formula is:

@@7Y - 10X@@
\end{lstlisting}

\subsection{Example o1-mini summary of Type 3 CRT output from gpt-4o (correct answer)}
\label{summ1}
\subsubsection{o1-mini input}
\begin{lstlisting}
7Y - 10X

Summarise the given formula into an AY - BX form, and put that into a json format, as shown below. Give only the json as your output, with no other text or formatting.  If it is not possible, then assign None to A and B. 
 
{
  'A': some_int,
  'B': some_int
}

 
Make sure that you don't make a mistake with negatives. If the answer is 10Y - 5X, then A is 10 and B is 5 (not negative 5).

\end{lstlisting}
\subsubsection{o1-mini response}
\begin{lstlisting}
RESPONSE:
{ 'A': 7, 'B': 10 }
\end{lstlisting}

\subsection{Example o1-mini summary of Type 3 CRT output from gpt-3.5-turbo (incorrect answer)}
\label{summ2}
\subsubsection{Turbo 3.5 answer to our example prompt}
\begin{lstlisting}

Let the initial area of light in the room be A and the final area of the room be 3A. We can represent the area of light in the room at any given time t as A(t) = A * 3^(t/5).

Given that it takes 7Y nanoseconds for the room to fill, we have A(7Y) = 3A.

To find the time when the room was 1/9 full, we need to solve for t when A(t) = A/9:

A(t) = A * 3^(t/5) = A/9
3^(t/5) = 1/9
t/5 = log3(1/9)
t = 5 * log3(1/9)

Therefore, the room was 1/9 full after t = 5 * log3(1/9) nanoseconds.

@@t = 5 * log3(1/9)@@
\end{lstlisting}

\subsubsection{o1-mini input}
\begin{lstlisting}
t = 5 * log3(1/9)

Summarise the given formula into an AY - BX form, and put that into a json format, as shown below. Give only the json as your output, with no other text or formatting.  If it is not possible, then assign None to A and B. 
 
{
  'A': some_int, 
  'B': some_int
}

Make sure that you don't make a mistake with negatives. If the answer is 10Y - 5X, then A is 10 and B is 5 (not negative 5).
\end{lstlisting}

\subsubsection{o1-mini response}
\begin{lstlisting}
{
  "A": null,
  "B": null
}
\end{lstlisting}

\end{document}